\documentclass[10pt, letterpaper]{article}

\usepackage[english]{babel}
\usepackage[utf8]{inputenc}
\usepackage[T1]{fontenc}
\usepackage{mathptmx} 
\usepackage{amssymb}
\usepackage[letterpaper, top=1in, bottom=1in, left=1.5in, right=1.5in]{geometry}

\usepackage{amsmath}
\usepackage{graphicx}
\usepackage{xcolor}
\usepackage{booktabs}
\usepackage{microtype} 
\usepackage[font=small, labelfont=bf, skip=6pt]{caption} 
\usepackage[colorlinks=true, allcolors=blue]{hyperref}
\usepackage{cleveref} 
\usepackage{multicol}
\usepackage{multirow}
\usepackage{listings}
\usepackage[style=ieee]{biblatex}
\usepackage{placeins}
\title{\textbf{\Large Reproducing Recurrent Transformers: The CoTFormer}}
\author{
  \textbf{Alberto Berni}\textsuperscript{1} \qquad \textbf{Bryan Vullo}\textsuperscript{1} \qquad \textbf{Aras Kavuncu}\textsuperscript{1} \\
  \setcounter{footnote}{0}
  \small\textsuperscript{1}School of Electronics and Computer Science, University of Southampton
}
\date{}

\begin{document}

\maketitle

\begin{abstract}
The CoTFormer architecture formalizes Chain-of-Thought as a form of \emph{recurrent latent computation}, preserving intermediate states as attendable representations to mimic explicit reasoning traces. In this work, we evaluate CoTFormer and its structural variants across perplexity and compute efficiency metrics. Furthermore, we extend evaluation to controlled algorithmic settings to determine whether this recurrent framework improves out-of-distribution generalisation on inductive reasoning tasks.
\end{abstract}

\section{Introduction}\label{sec:intro} 

Transformers replaced recurrence with parallel self-attention, enabling scalable language modelling but weakening the architectural bias toward iterative computation \cite{vaswani2017attention}. This tension is central to neural algorithmic reasoning: earlier recurrent models (i.e., Neural GPUs) could generalise from short training instances to longer algorithmic inputs through repeated parameter sharing \cite{kaiser2016neural}. Block Universal Transformers (BUT) later reintroduced this idea by recurrently applying a fixed, weight-tied block of multiple Transformer layers to token sequences, establishing a baseline configuration for weight-tied architectural recurrence \cite{dehghani2019universal}. \\

The CoTFormer architecture formalizes this connection by routing hidden state representations through the same sequence of weight-tied layers within a single forward pass, avoiding the parameter footprint of deep, un-tied layer stacks. Crucially, CoTFormer preserves previous intermediate states as attendable representations, mimicking explicit thought tokens attending to prior reasoning steps \cite{cotformer}. Prior evaluations focus on perplexity and compute efficiency; however, recent work indicates that looped models can improve performance on reasoning-heavy tasks despite exhibiting worse perplexity and memorisation than iso-FLOP baselines calibrated to equal floating-point compute budgets \cite{saunshi2025reasoning}. This motivates evaluating CoTFormer as a recurrent architecture with an inductive bias for iterative reasoning. We reproduce and evaluate the main claims of CoTFormer and its variants, testing if its recurrent structure improves controlled algorithmic generalisation.
\section{Methodology}\label{sec:methodology}
Reproduction of this work\footnote{Recorded on \href{https://github.com/COMP6258-Reproducibility-Challenge/CoTFormer}{GitHub}, forked from the \href{https://github.com/epfml/CoTFormer}{original}.} was performed on L4 GPUs, requiring specific adaptations to the original codebase, designed for A100 GPUs. To reconcile this divergence, we applied the following: \textbf{i. Minor Adaptations} Reduced vRAM budget required lowering batch size and proportionally increasing gradient accumulation steps, retaining the original batch size (128). Evaluation of MAC metrics was estimated in closed-form due to these constraints. \textbf{ii. LN-CoTFormer} We initially trained this variant to 40k steps for \Cref{tab:table2-repro}, resuming to 60k steps for further evaluation. This invalidated the previous learning-rate scheduler, requiring use of a different one and causing the 0.4 perplexity divergence observed in \Cref{tab:section5-pair}. \textbf{iii. ADM Distributed Training} The original repository bypassed RNG state checkpointing (required by the ADM). Replacing the NCCL backend with Gloo guaranteed RNG state restoration while confounding training dynamics (\cref{sec:ADM}). For other experiments, a separate branch fixed NCCL's deadlock and RNG state checkpointing issues.

\section{Reproduction}
\paragraph{Base CotFormer} 
Base CoTFormer models are reproducible; we notice ``\emph{benign spikes}'' during training, characterized as temporary, non-fatal surges in gradient norms or training perplexity that recover quickly without causing optimization divergence or permanent destabilization, later explored in~\cref{subsec:phase-change}.

\begin{figure}[!ht]
    \centering
    \begin{minipage}{0.31\textwidth}
        \centering
        \textbf{\scriptsize(a)} \\
        \includegraphics[width=\linewidth]{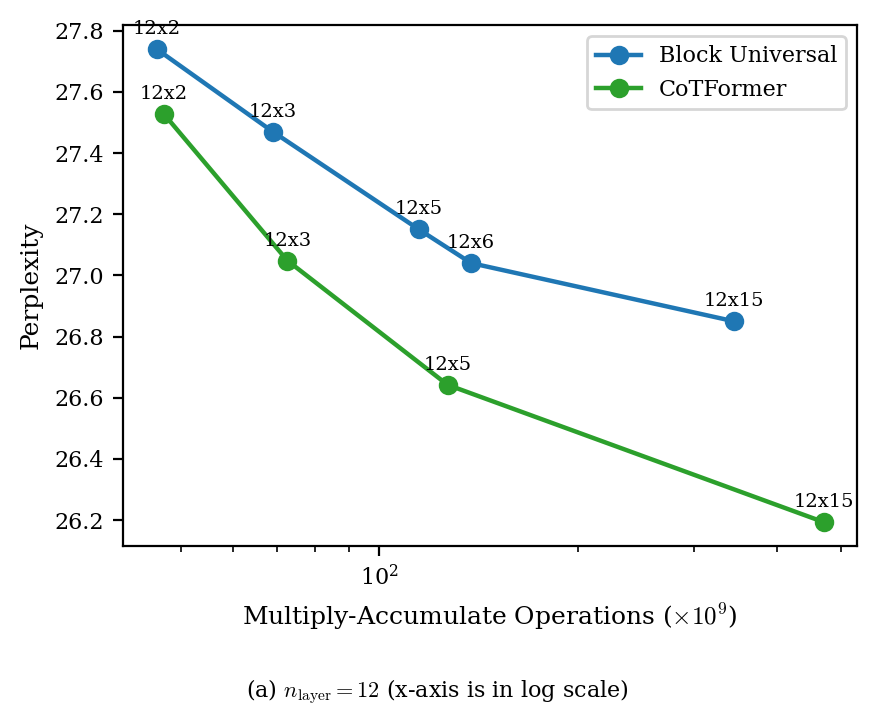}
    \end{minipage}\hfill
    \begin{minipage}{0.31\textwidth}
        \centering
        \textbf{\scriptsize(b)} \\
        \includegraphics[width=\linewidth]{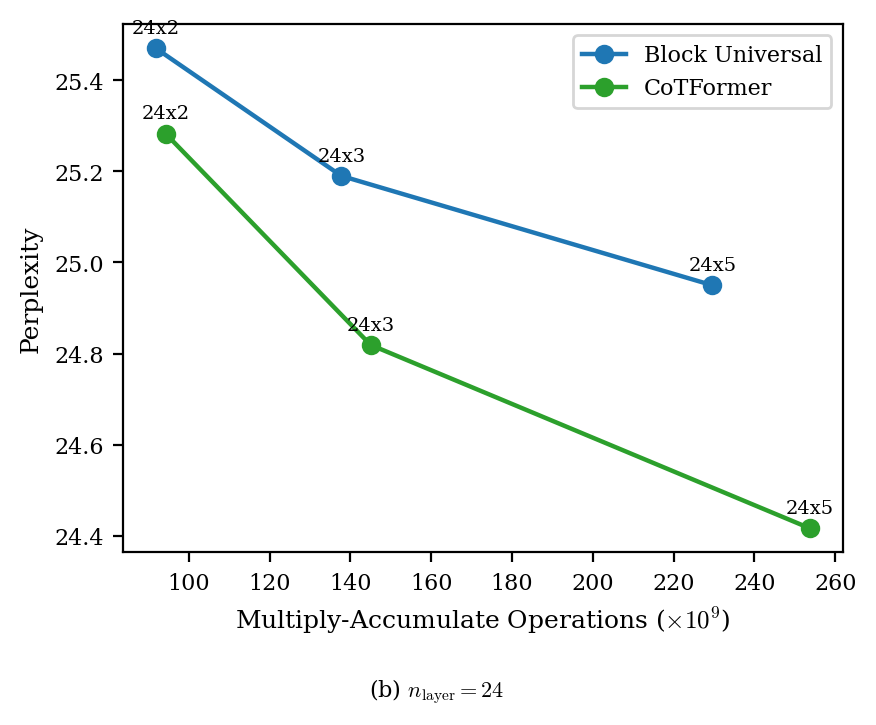}
    \end{minipage}\hfill
    \begin{minipage}{0.31\textwidth}
        \centering
        \textbf{\scriptsize(c)} \\
        \includegraphics[width=\linewidth]{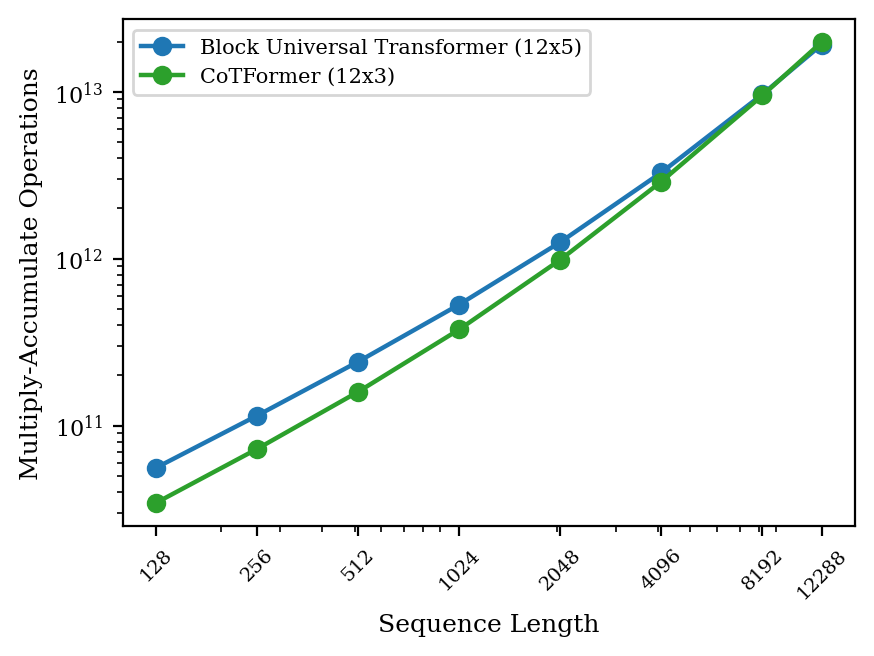}
    \end{minipage}
    \caption{\textbf{(a)-(b)} Reproduced comparison of BUT (original results) and CotFormer pareto curves; \textbf{(c)} Reproduced comparison of compute intensity when performance is comparable (Paper fig. 1a, 1b, 2).}
    \label{fig:2ab-and-3}
\end{figure}

\begin{table}[!ht]
\centering
\small
\setlength{\tabcolsep}{3.8pt} 
\begin{tabular}{lc ccc ccc ccc}
\toprule
 & & \multicolumn{3}{c}{Theirs (SEM)} & \multicolumn{3}{c}{Ours ($\text{CI}_{95}$)} & \multicolumn{3}{c}{$\Delta$ ($\text{Ours} - \text{Theirs}$)} \\
\cmidrule(lr){3-5} \cmidrule(lr){6-8} \cmidrule(lr){9-11}
Model & $n_\text{layer}$ & 2 & 3 & 5 & 2 & 3 & 5 & 2 & 3 & 5 \\
\midrule
CoTFormer & 12 & 27.55 & 27.07 & 26.64 & 27.53 & 27.05 & 26.64 & $-0.02$ & $-0.02$ & $\phantom{-}0.00$ \\
CoTFormer & 24 & 25.28 & 24.85 & 24.48 & 25.28 & 24.82 & 24.42 & $\phantom{-}0.00$ & $-0.03$ & $-0.06$ \\
\bottomrule
\end{tabular}
\caption{Perplexity comparison across varying repeat counts ($n_\text{repeat}$). \textbf{Theirs} are the original  results, reported with  Standard Error of the Mean (SEM) over 3 seeds. \textbf{Ours} represents a single-seed evaluation showing per-batch $95\%$ confidence intervals. $\Delta$ values fall within these boundaries, confirming exact reproducibility.}
\label{tab:table1}
\end{table}

\paragraph{CoTFormer Variants} The original CoTFormer architecture is extended by three compounding variants: \textbf{i.}~a standard CoTFormer that reserves first and last layers for embedding translation alone \textbf{ii.}~LN-CoTFormer adds a post-repeat LayerNorm to stabilise the recursive residual stream, and \textbf{iii.}~Adaptive Depth Module (ADM), a dynamic token routing mechanism that predicts whether individual tokens require further recurrent iterations or can exit early to optimize inference compute. We reproduce the variants and evaluate their structural hypotheses.
\begin{table}[!ht]
\centering
\small
\setlength{\tabcolsep}{1.0pt}
\begin{tabular}{llcccc}
\toprule
\textbf{Variant} & \textbf{Layers (Effective)} & \textbf{Paper PPL (SEM)} & \textbf{Ours PPL} & $\Delta$ & \textbf{95\% CI (Ours)} \\ 
\midrule
CoTFormer & $24\times5$ (120) & 24.48 (0.03) & 24.42 & $-0.06$ & [23.84, 25.01] \\
CoTFormer + Reserved & $24, 2\rightarrow21\times5\rightarrow1$ (108) & 24.51 (0.01) & 24.64 & $+0.13$ & [24.25, 25.03] \\
LN-CoTFormer ($40$k) & $24, 2\rightarrow21\times5\rightarrow1$ (108) & 24.11 (0.03) & 24.13 & $+0.02$ & [23.75, 24.51] \\ 
\bottomrule
\end{tabular}
\caption{Reproduction of the CoTFormer's Table 2 results at \texttt{40k} training steps.}
\label{tab:table2-repro}
\end{table}
\paragraph{Reserved Layers \& LN-CoTFormer}\label{sec:RLLN}
CoTFormer with Reserved Layers reproduces its original results within statistical bounds (\Cref{tab:table2-repro}). This change alone\footnote{Isolating recurrent reasoning in a central block by reserving the first and last layers strictly for embedding translation.} \emph{isn't} more parameter-efficient than the baseline Transformer (48L) nor the CoTFormer ($24\times5$) on perplexity alone. The 24-layer model effectively executes a 108-layer forward pass, showing it can \emph{approximate} deeper standard model capacity, paying a cost in sheer performance. \textbf{LayerNorm:} Authors justify this variant claiming ``\emph{it is important to maintain a consistent input’s scale (across repeats)}''; the insight follows from the original architecture's \cite{vaswani2017attention} and remains empirically valid given the 0.51 PPL improvement from the baseline CoTFomer. Crucially, the variant outperforms the 48-layer Transformer baseline, a much more widely adopted architecture, by 0.29 PPL. While the result shows recursive computation (similarly to parallel counterparts) benefits from explicit feature scaling to prevent representation drift, it fails to explain \emph{why} this is the case.

\begin{figure}[!ht]
    \centering
    \begin{minipage}{0.24\textwidth}
        \centering
        \textbf{\scriptsize(a)} \\
        \includegraphics[width=\linewidth]{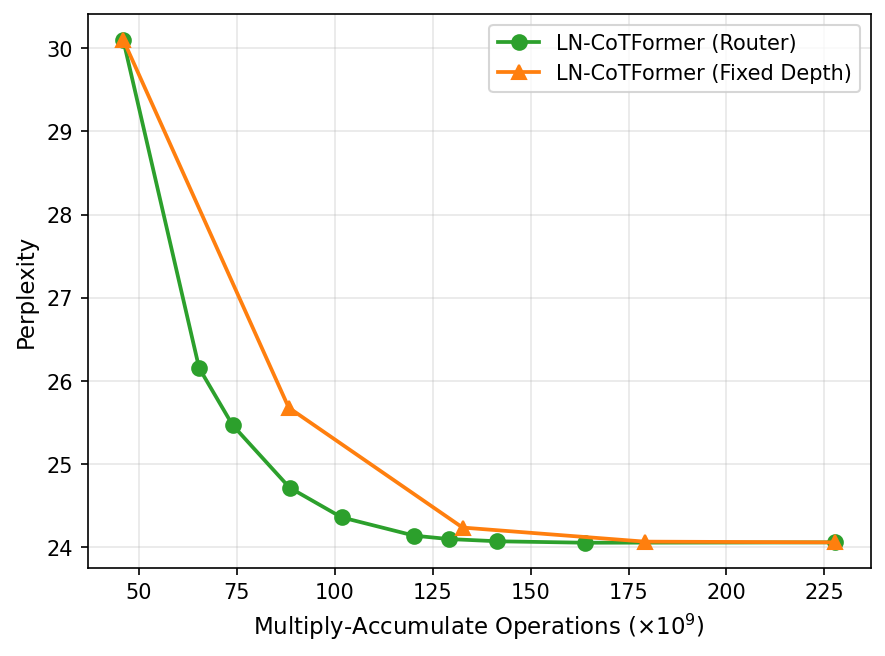}
    \end{minipage}\hfill
    \begin{minipage}{0.24\textwidth}
        \centering
        \textbf{\scriptsize(b)} \\
        \includegraphics[width=\linewidth]{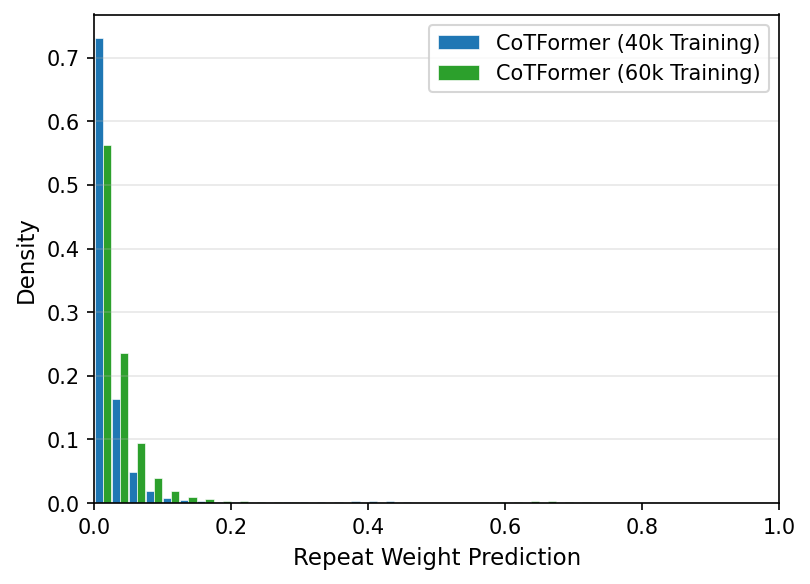}
    \end{minipage}\hfill
    \begin{minipage}{0.24\textwidth}
        \centering
        \textbf{\scriptsize(c)} \\
        \includegraphics[width=\linewidth]{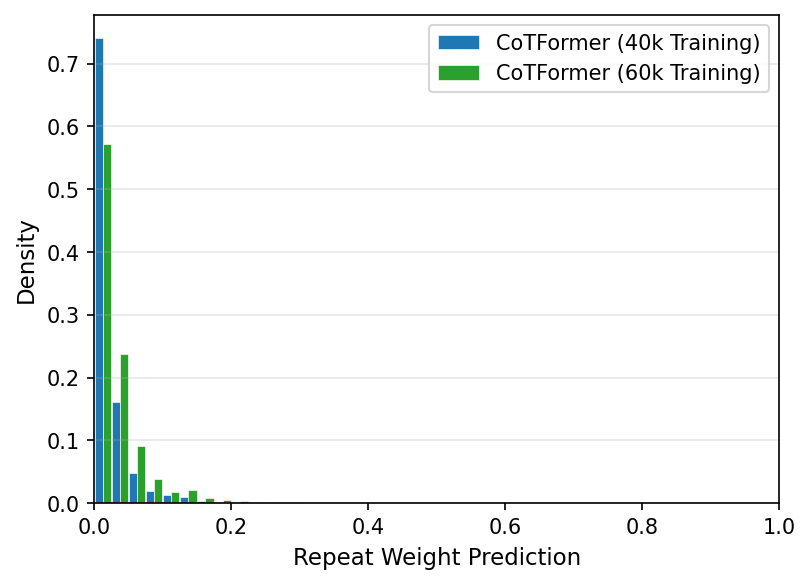}
    \end{minipage}\hfill
    \begin{minipage}{0.24\textwidth}
        \centering
        \textbf{\scriptsize(d)} \\
        \includegraphics[width=\linewidth]{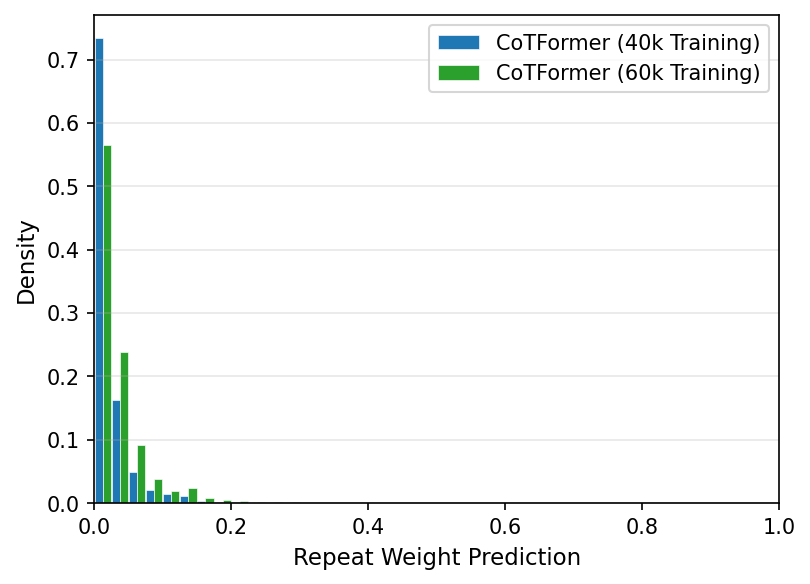}
    \end{minipage}
    \caption{\textbf{(a)} Reproduced Pareto curve of compute vs perplexity; \textbf{(b)-(d)} Comparison of router weight extraction methods (raw, per-index cascade, and per-token cascade) at 40k and 60k training steps.}
    \label{fig:adm-evaluation}
\end{figure}

\paragraph{ADM CoTFormer}\label{sec:ADM} By employing a ``\emph{Mixture of Repeats}'' strategy with randomized capacities, the ADM trains a routing mechanism to decide whether a token requires further recurrent processing~\cite{cotformer}. While the original Figure 4 (\cref{fig:adm-evaluation}\textbf{(a)}) can be replicated, the core claim that ``\emph{the ADM can efficiently route computation based on token difficulty}'' cannot be confirmed, as it lacks a clear definition of efficiency: showing the ADM outperforms its fixed-compute-budget copy is hardly a proof of this. 

\paragraph{Irreproducible claims} Authors state the adaptive model ``\emph{would benefit more from longer training}'', proposing extended training budgets. Using Figure \cite{cotformer}, they argue the model ``\emph{increasingly utilises deeper repeats over time}''. The claims are unrelated: increased usage of final repeats only proves the model allocates more compute; verifying a predictive benefit requires controlled evaluation at matched inference compute (i.e. \cref{fig:2ab-and-3}). Further, the authors' extraction script contains an inaccuracy in how it orders extracted weights (\texttt{get\_router\_weights.py}, 129-131), whereby the intended ``\emph{final repeat}'' becomes a cascaded minimum across repeats. We replicate this exactly in \cref{fig:adm-evaluation}~\textbf{(c)}, further evaluating two different extraction methods (\cref{fig:adm-evaluation}): \textbf{(b)}~the raw router weights, and \textbf{(d)}~corrected per-token cascade. Across all histograms, densities decay drastically compared to the original figure: ADM final repeat usage is lower than claimed. This divergence is confounded by back-end divergences (\cref{sec:methodology}); we can qualitatively confirm a \emph{marginal} increase in final repeat usage in longer training, but cannot conclude whether the claimed distribution is genuine or a by-product of runtime constraints.

\begin{table}[!ht]
\centering
\small
\begin{tabular}{lccc}
\toprule
\textbf{Model} & \textbf{Paper PPL} & \textbf{Ours PPL (SEM)} & $\Delta$ \\ \midrule
LN CoTFormer 60k & 23.19 & 23.59 (0.01) & +0.40 \\
ADM 60k (full compute) & 23.83 & 24.06 (0.01) & +0.23 \\ \midrule
Adaptive cost & +0.64 & +0.47 & N/A \\ \bottomrule
\end{tabular}
\caption{Reproduced CoTFormer Section 5 PPL \cite{cotformer} comparison at 60k training steps.}
\label{tab:section5-pair}
\end{table}
\section{Extensions}
\subsection{Phase change analysis} \label{subsec:phase-change}

The reproduction runs showed transient spikes in gradient norm and loss. We use
``phase change'' descriptively to refer to a sharp training-time reorganisation
in the model's diagnostic statistics, rather than as evidence of a discrete
mechanistic transition. We therefore tested whether these spikes coincided with
changes in how CoTFormer uses cached repeat states.

To do this, we tracked two complementary properties of repeat-state attention:
how much attention mass the final repeated block assigns to each cached repeat
state, and how concentrated that attention is within each repeat. During
training, the model can use PyTorch's flash attention implementation for speed.
During evaluation, the implementation falls back to explicit attention so that
the post-softmax attention probabilities can be stored and analysed.

We now define the diagnostics used in this analysis. Let \(B\), \(H\), \(Q\),
and \(R\) denote batch size, number of heads, sequence length, and number of
middle-block repeats. In evaluation mode, we store the post-softmax attention
from the final middle-block attention layer and reshape it as
\[
    A_{b,h,q,r,k}
    \in
    \mathbb{R}^{B \times H \times Q \times R \times Q},
\]
where \(q\) indexes the query token, \(r\) the cached repeat state, and \(k\)
the key token inside that repeat. For fixed \(b,h,q\),
\[
    \sum_{r=1}^{R}\sum_{k=1}^{Q} A_{b,h,q,r,k}=1.
\]

The repeat attention budget measures how much attention mass is assigned to
each cached repeat:
\[
    m_{b,h,q,r} = \sum_{k=1}^{Q} A_{b,h,q,r,k},
    \qquad
    \bar{m}_{r}
    =
    \frac{1}{BHQ}
    \sum_{b,h,q} m_{b,h,q,r}.
\]
We also compute entropy over repeat states,
\[
    H^{\mathrm{repeat}}_{b,h,q}
    =
    -\sum_{r=1}^{R}
    m_{b,h,q,r}
    \log(m_{b,h,q,r}+\epsilon),
    \qquad
    \bar{H}^{\mathrm{repeat}}
    =
    \frac{1}{BHQ}
    \sum_{b,h,q}
    H^{\mathrm{repeat}}_{b,h,q},
\]
which is low when attention is concentrated on a small number of repeats.

Finally, to distinguish repeat-level allocation from token-level focus inside a
repeat, we normalise within each repeat:
\[
    P_{b,h,q,r,k}
    =
    \frac{A_{b,h,q,r,k}}{m_{b,h,q,r}+\epsilon},
\]
and compute within-repeat entropy
\[
    H^{\mathrm{within}}_{b,h,q,r}
    =
    -\sum_{k=1}^{Q}
    P_{b,h,q,r,k}
    \log(P_{b,h,q,r,k}+\epsilon),
    \qquad
    \bar{H}^{\mathrm{within}}_{r}
    =
    \frac{1}{BHQ}
    \sum_{b,h,q}
    H^{\mathrm{within}}_{b,h,q,r}.
\]
Together, \(\bar{m}_{r}\), \(\bar{H}^{\mathrm{repeat}}\), and
\(\bar{H}^{\mathrm{within}}_{r}\) track whether attention shifts between repeat
states and whether attention within a selected repeat becomes more concentrated
or more diffuse.

\begin{figure}[!h]
    \centering
    \begin{minipage}{0.23\textwidth}
        \centering
        \textbf{\scriptsize(a)}\\
        \includegraphics[width=\textwidth]{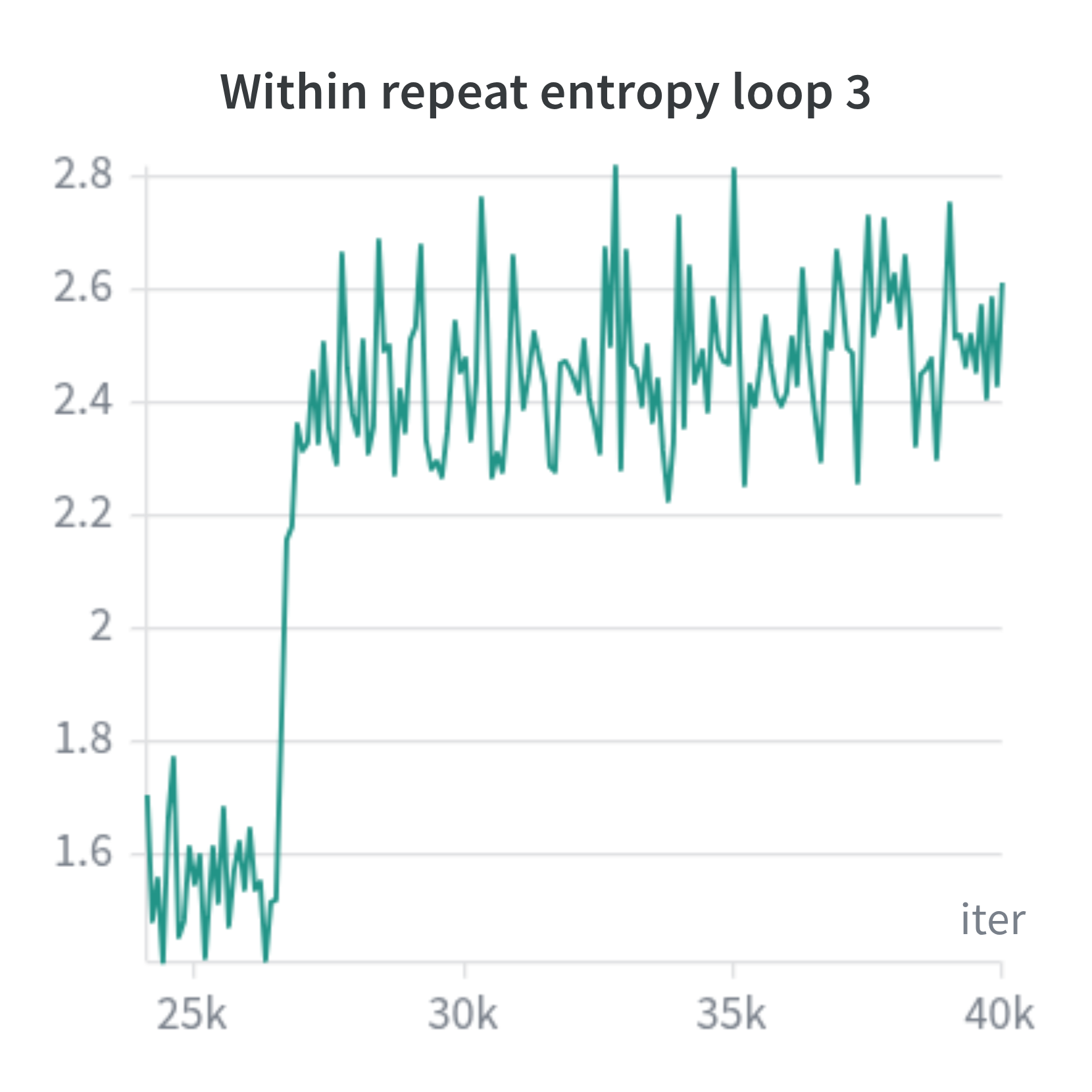}
    \end{minipage}\hfill
    \begin{minipage}{0.23\textwidth}
        \centering
        \textbf{\scriptsize(b)}\\
        \includegraphics[width=\textwidth]{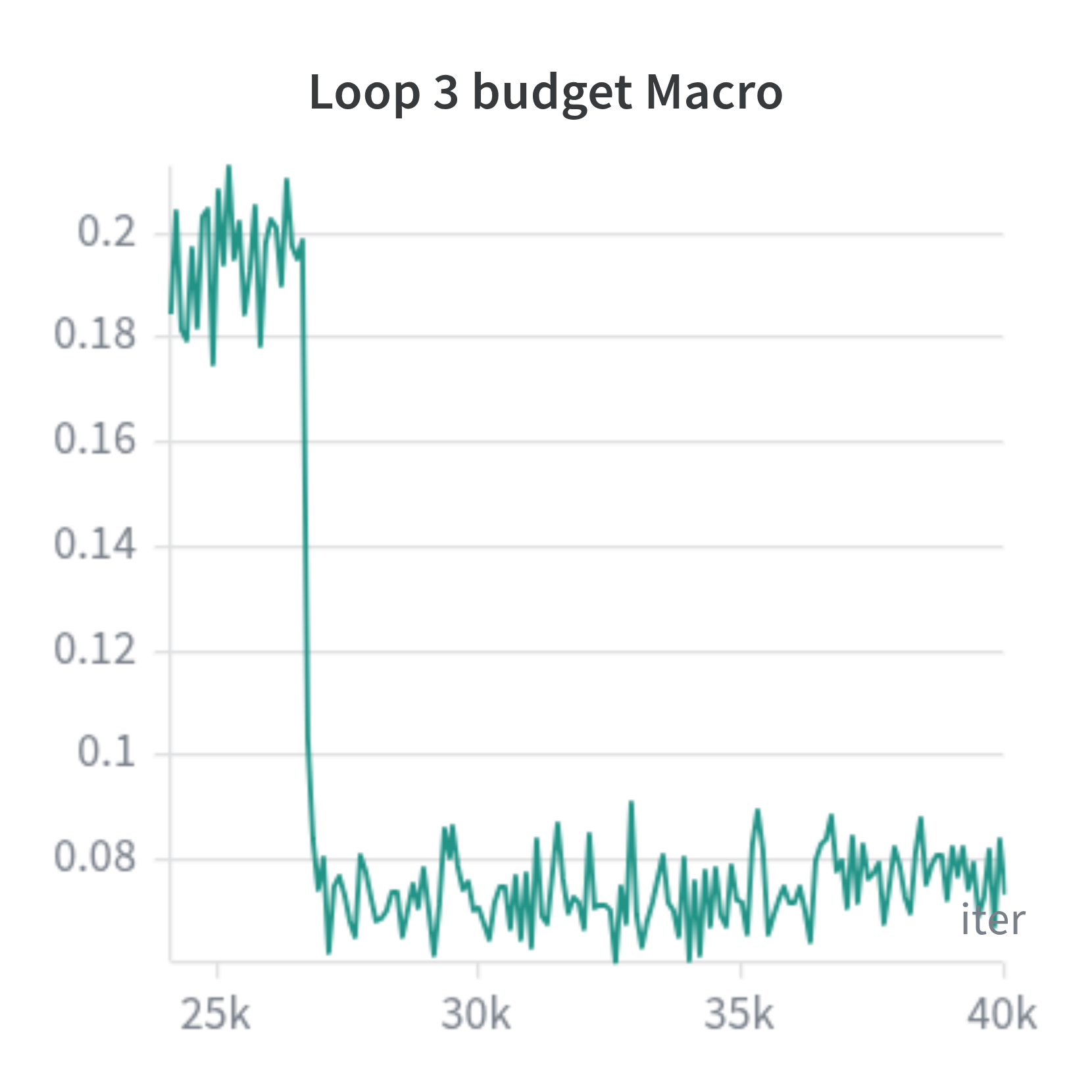}
    \end{minipage}\hfill
    \begin{minipage}{0.23\textwidth}
        \centering
        \textbf{\scriptsize(c)}\\
        \includegraphics[width=\textwidth]{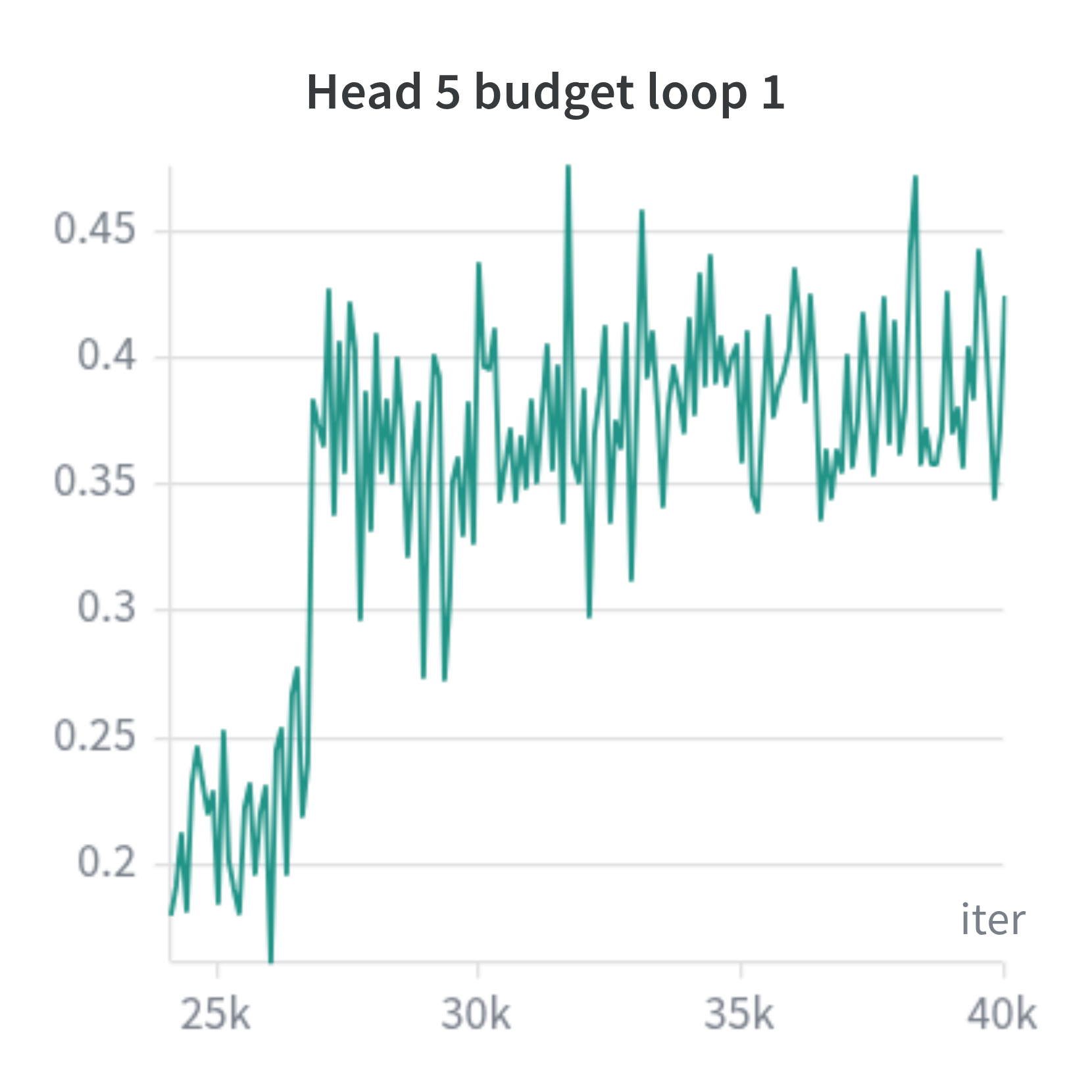}
    \end{minipage}\hfill
    \begin{minipage}{0.23\textwidth}
        \centering
        \textbf{\scriptsize(d)}\\
        \includegraphics[width=\textwidth]{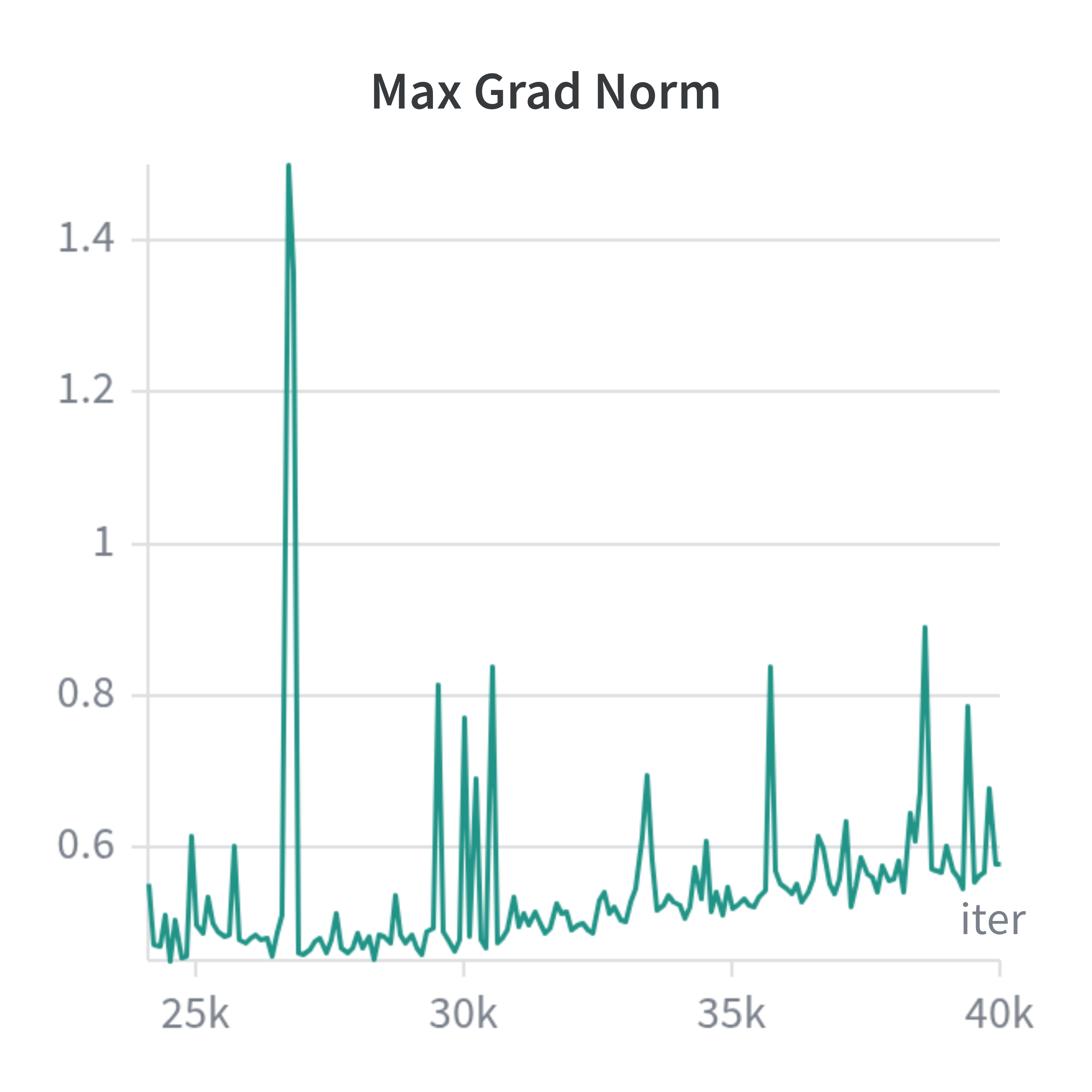}
    \end{minipage}
    \caption{Repeat-state attention reorganisation during OWT2 training of a $2\rightarrow24\times5\rightarrow1$ CoTFormer.}
    \label{fig:combined-cascade-plots}
\end{figure}

Armed with these metrics, during the training of a CoTFormer with 2 begin reserve layers 24 middle layers repeating 5 times and 1 end layer we notice a significant shift in attention distribution across CoTFormer repeat states during training (\cref{fig:combined-cascade-plots}). Around 27k training steps, attention mass redistributes rapidly across different repeat budgets, accompanied by temporary, self-recovering surges in gradient norm (benign spikes) and a substantial increase in within-repeat attention entropy for certain repeats. We interpret this as evidence cosistent with repeat-state specialisation: middle blocks may learn non-uniform uses of latent repeat states. Establishing if these correspond to algorithmic recurrent reasoning requires additional experiments, which we leave to future work.

\FloatBarrier
\subsection{Shifted Start}
To answer our research question (\cref{sec:intro}), we adapt the shifted-start counting task from Chang and Bisk~\cite{chang2025counting}, which evaluates abstract sequence tracking by forcing the model to increment integers from an arbitrary non-standard initialization point (e.g., starting at 47 instead of 1) to disrupt simple positional memory shortcuts. We train on sequences up to length 25 and test on 200. We extend the limited training split with variable-length examples below the maximum training length. Early high-width runs were unstable and prone to overfitting, so we reduced width to 128 and used 4 heads while keeping the expanded dataset. Across BUT, CoTFormer and Transformer baselines, OOD accuracy remained capped around 0.26--0.27. Increasing loop depth did not improve extrapolation under teacher-forced sequence prediction, suggesting that models may rely on positional shortcuts rather than an iterative counting procedure. The strongest single-seed result came from the reserved-layer CoTFormer, although the effect is seed-sensitive.\\
\begin{table}[!ht]
\centering
\small
\setlength{\tabcolsep}{6.0pt}
\begin{tabular}{lcccc}
\toprule
Model Configuration & Seeds & Mean OOD Acc. & Std. Dev. & Max OOD Acc. \\
\midrule
CoTFormer: \(1 \to 1{\times}3 \to 1\) & 4 & 0.2680 & 0.0091 & 0.2809 \\
CoTFormer: \(1{\times}8\)             & 2 & 0.2597 & 0.0030 & 0.2618 \\
CoTFormer: \(1{\times}4\)             & 4 & 0.2617 & 0.0039 & 0.2663 \\
Base Transformer: \(4{\times}1\)      & 4 & 0.2674 & 0.0030 & 0.2703 \\
Base Transformer: \(1{\times}1\)      & 4 & 0.2348 & 0.0034 & 0.2381 \\
\bottomrule
\end{tabular}
\caption{Shifted-start out-of-distribution (OOD) accuracy summary. Formatted to four decimal places to maintain the precision limits mandated by p-hop experiments.}
\label{tab:ood_acc_summary}
\end{table}
\subsection{P-hop induction}
\paragraph{Dataset construction.}
For p-hop induction, we used the constructive data generator in
\texttt{p-hop-induction/generate.py} and \texttt{p-hop-induction/phop\_data.py}.
The task was fixed-hop final-answer p-hop induction with \(p=32\), sequence
length \(n=256\), and alphabet size \(|\Sigma|=4\), where
\(\Sigma=\{a,b,c,d\}\). Since the hop count was fixed, no explicit hop token was
included in the input.

Each example consists of an input sequence
\[
    x = (x_1,\ldots,x_n), \qquad x_i \in \Sigma,
\]
and a label sequence
\[
    y = (-1,\ldots,-1,a^\star),
\]
where all positions except the final query position are masked with
\(-1\). Thus, the model is trained as a sequence model but supervised only on
the final p-hop answer \(a^\star\).

The p-hop target is computed by starting from the final token \(x_n\). At each
hop, the procedure finds the nearest previous occurrence of the current query
symbol and moves to the token immediately after that occurrence. Formally, if
the current state at hop \(t\) is \((i_t, x_{i_t})\), then
\[
    j_t = \max\{j < i_t : x_j = x_{i_t}\},
    \qquad
    i_{t+1}=j_t+1.
\]
After \(p\) hops, the target is
\[
    a^\star = x_{i_p}.
\]

The constructive generator first plants a valid
p-hop chain and then fills the remaining sequence positions subject to
constraints that preserve the planted nearest-previous-occurrence traversal.
This ensures that every generated example has a valid \(p\)-hop answer in
\(\Sigma\). We generated the constructive splits
\texttt{train\_constructive}, \texttt{val\_constructive}, and
\texttt{test\_constructive}, containing \(2{,}000{,}000\), \(50{,}000\), and
\(50{,}000\) examples respectively. The base random seed was \(0\), with split
seeds incremented by split order.
\begin{table}[!ht]
\centering
\small
\begin{tabular}{lrrrr}
\toprule
Model & Eff. depth & Step & Val acc & Test acc \\
\midrule
Base Transformer \(1{\times}1\)        & 1 & 240k & 0.4659 & 0.4676 \\
Base Transformer \(6{\times}1\)        & 6 & 300k & 0.6121 & 0.6133 \\
CoTFormer \(1{\times}4\)               & 4 & 248k & 0.5935 & 0.6005 \\
BUT \(1{\times}6\)                     & 6 & 250k & 0.6582 & 0.6597 \\
CoTFormer \(1{\times}6\)               & 6 & 300k & 0.6627 & 0.6619 \\
CoTFormer \(1 \to 1{\times}4 \to 1\)   & 6 & 272k & 0.6605 & 0.6585 \\
\bottomrule
\end{tabular}
\caption{Single-seed p-hop induction results on constructive \(p=32\), \(n=256\), \(|\Sigma|=4\).}
\label{tab:phop_results}
\end{table}
\paragraph{P-hop induction results.}

P-hop induction can be viewed as a repeated retrieval problem related to
induction-head behaviour: each hop requires finding a previous occurrence of a
query symbol and reading the following token. Since \(p=32\), success requires
composing this operation many times rather than performing a single local copy.
We use the task as a behavioural probe for iterative retrieval, without claiming
a mechanistic identification of induction heads.

Unlike shifted-start inductive counting, the p-hop task benefits from repeated computation. In shifted-start counting, increasing loop depth did not improve OOD extrapolation, suggesting that recurrence over layers did not induce a true counter-like procedure, while on p-hop induction, looped models clearly outperform shallow Transformer baselines on this run. The \(1{\times}1\) Transformer reaches only \(0.4676\) test accuracy, while \(1{\times}6\) BUT and CoTFormer reach roughly \(0.66\). This supports a task-dependent interpretation: looped computation appears useful when the target algorithm resembles repeated retrieval or pointer chasing, but it does not automatically provide the successor-state mechanism required for inductive counting.\\

The CoTFormer-specific advantage is modest. CoTFormer \(1{\times}6\) slightly outperforms BUT, and the reserved-middle variant \(1 \to 1{\times}4 \to 1\) performs similarly. These results support the usefulness of repeated latent computation for p-hop induction, but do not isolate a large benefit of CoTFormer's cache mechanism over vanilla block looping.
\section{Conclusion}
We largely reproduce the main CoTFormer perplexity results, including the benefit of post-repeat LayerNorm, but find the adaptive-depth claims less clean: ADM preserves a compute--perplexity tradeoff, while its reported deep-repeat usage is sensitive to extraction method and training confounders. Our extensions suggest a narrower interpretation of CoTFormer’s recurrent bias. Repeated latent computation helps on p-hop induction, where the target resembles iterative retrieval, but does not by itself induce robust OOD counting on shifted-start. Thus, CoTFormer appears promising as a compute-sharing architecture with task-dependent algorithmic benefits, rather than a general mechanism for latent chain-of-thought reasoning. Stronger claims require matched-compute evaluations, more seeds, and controlled tasks isolating when recurrence learns an actual iterative procedure.
\section*{Author Contributions}
AB and BV conducted the reproducibility component of the project, including
reproduction and analysis of the original CoTFormer results. The extensions in
Section 4 were designed, implemented, and evaluated by TAK, including the
phase-change analysis, shifted-start counting task, and p-hop induction
experiments. All authors discussed the results and reviewed the final manuscript.
\printbibliography[heading=bibnumbered, title={References}]

@inproceedings{cotformer,
    title={CoTFormer: A Chain-of-Thought Driven Architecture with Budget-Adaptive Computation Cost at Inference},
    author={Amirkeivan Mohtashami and Matteo Pagliardini and Martin Jaggi},
    booktitle={The Thirteenth International Conference on Representation Learning (ICLR)},
    year={2025},
    url={https://openreview.net/forum?id=7igPXQFupX}
}

@inproceedings{dehghani2019universal,
    title={Universal Transformers},
    author={Mostafa Dehghani and Stephan Gouws and Oriol Vinyals and Jakob Uszkoreit and {\L}ukasz Kaiser},
    booktitle={International Conference on Learning Representations (ICLR)},
    year={2019}
}

@inproceedings{vaswani2017attention,
    title={Attention Is All You Need},
    author={Vaswani and Shazeer and Parmar and Uszkoreit and Jones and Gomez and Kaiser and Polosukhin},
    booktitle={Advances in Neural Information Processing Systems (NeurIPS)},
    year={2017}
}

@inproceedings{kaiser2016neural,
    title={Neural GPUs Learn Algorithms},
    author={Kaiser and Sutskever},
    booktitle={International Conference on Learning Representations (ICLR)},
    year={2016},
    note={arXiv:1511.08228}
}

@inproceedings{saunshi2025reasoning,
    title={Reasoning with Latent Thoughts: On the Power of Looped Transformers},
    author={Saunshi and Dikkala and Li and Kumar and Reddi},
    booktitle={International Conference on Learning Representations (ICLR)},
    year={2025},
    note={arXiv:2502.17416}
}

@inproceedings{chang2025counting,
    title={Language Models Need Inductive Biases to Count Inductively},
    author={Chang and Bisk},
    booktitle={International Conference on Learning Representations (ICLR)},
    year={2025}
}

\end{document}